\documentclass[11pt,a4paper]{article}
\usepackage[hyperref]{emnlp2020}
\usepackage{times}
\usepackage{latexsym}
\usepackage{graphicx}
\usepackage{float}
\graphicspath{ {emnlp2020-templates/images/} }

\usepackage{microtype}

\usepackage{color}

\aclfinalcopy 

\newcommand\blfootnote[1]{%
  \begingroup
  \renewcommand\thefootnote{}\footnote{#1}%
  \addtocounter{footnote}{-1}%
  \endgroup
}

\title{CXP949 at WNUT-2020 Task 2: Extracting Informative COVID-19 Tweets - RoBERTa Ensembles and The Continued Relevance of Handcrafted Features} 
\author{Calum Perrio \\
  School of Computer Science \\
  University of Birmingham \\
  United Kingdom\\
  \texttt{\small cperrio2015@gmail.com} \\\And
Harish Tayyar Madabushi \\
  School of Computer Science \\
  University of Birmingham\\
  United Kingdom \\
  \texttt{ \small harish@harishtayyarmadabushi.com} \\
  }

\date{September 2020}

\begin{document}
\maketitle
\begin{abstract}

This paper presents our submission to Task 2 of the Workshop on Noisy User-generated Text. We explore improving the performance of a pre-trained transformer-based language model fine-tuned for text classification through an ensemble implementation that makes use of corpus level information and a handcrafted feature. We test the effectiveness of including the aforementioned features in accommodating the challenges of a noisy data set centred on a specific subject outside the remit of the pre-training data. We show that inclusion of additional features can improve classification results and achieve a score within 2 points of the top performing team.

\end{abstract}

\section{Introduction}
\blfootnote{
    \hspace{-0.65cm}
    \vspace{-0.65cm}
    Accepted for publication at the 6th W-NUT, 2020.
    }
Identification of informative tweets in relation of coronavirus presents a text classification problem. Pre-trained bidirectional transformer-based models such as such as BERT~\cite{BERT} and RoBERTa~\cite{RoBERTa} have proven to be extremely successful on text classification tasks; the process of pre-training on a large corpora enables the generation of effective contextual embeddings during fine-tuning, which can be leveraged on the classification task through the addition of an output layer \citep{BERT}. The progression of the state-of-the-art that these models have facilitated is clearly demonstrable on performance against the General Language Understanding Evaluation (GLUE) benchmark \citep{GLUE}; a collection of Natural Language Understanding tasks with an associated online platform for evaluation and analysis.

The corpora utilised during pre-training of transformer-based models typically consists of documents written in formal English. Moreover, this data is highly unlikely to contain references to the coronavirus pandemic which has only just occurred. Therefore, the noise inherent in social media data and subject specificity to coronavirus in the current data set \citep{covid19tweet}, present challenges to conducting text classification with a pre-trained transformer-based model alone. To this regard, supplementary information may be useful to improving performance; corpus level information may potentially capture notions of relevance to words which fall outside the pre-trained vocabulary, and handcrafted features may show additional distinctions between classes. In this paper, we describe the development of a system with includes such features through the use of an ensemble, and which in the final submission to the evaluation stage achieved an $F_1$ Score of 0.8910. 

The rest of this paper is organised as follows, firstly a discussion of related work is presented at Section~\ref{sec:relatedwork}. This is followed by a description of the methodological approach at Section~\ref{sec:meth}. We present our results and analysis at Section~\ref{sec:results} and present our conclusions in Section~\ref{sec:conclusion}. 

\section{Related Work}
\label{sec:relatedwork}

As the 6th Workshop of Noisy User-generated Text presents the first time Task 2 has been made available, we consider research utilising pre-trained bidirectional transformer-based models from similar tasks to better enable informed decision making. 

\subsection{Related Tasks}

SemEval-2020 Task 12: Multilingual Offensive Language Identification in Social Media \citep{semeval2020}, presented at sub-task A a similar binary classification problem to the present task. Many of the highest performing teams on this sub-task made use of ``contextualised BERT-style Transformers'' \citep{semeval2020}. Sub-task A specifically comprised of identifying whether a tweet presented content that contained inappropriate language, threats or insults, or was neither offensive nor profane \citep{semeval2020}. Additionally, the sub-task was split based on the language of the tweet text, and to this regard we only consider relevant systems on the English data set, which present the highest degree of correlation to Task 2.  

\citet{semeval2020wiedemann2020uhhlt} achieved first place on SemEval-2020 Task 12 sub-task A utilising an ensemble of RoBERTa~\citep{RoBERTa}. Additionally, the authors leveraged the Masked Language Model pre-training objective of RoBERTa and further pre-trained on an \textit{in-domain} data set. Domain specific pre-training is observed to be a method that is established as improving later results of supervised task-specific fine-tuning. Further pre-training is equally explored by \citet{sotudeh2020guir} in their submission on SemEval-2020 Task 12 sub-task A that utilised BERT and achieved 4\textsuperscript{th}place.  

\citet{lim2020uob} presented an ensemble model of BERT and TF-IDF. They hypothesise that corpus level count information captured by TF-IDF can boost the performance of BERT, and the authors achieve a result within 2 points of the top scoring team. Moreover, they note this performance was achieved using only 10\% of the available training data due to physical constraints. We implement a similar method of incorporating TF-IDF features described at Section~\ref{sec:meth}, but differentiate from this work through our ability to incorporate the entire training data.

\subsection{User-generated Content}
\label{sec:rl:ugc}
A key aspect of the present task to consider is the nature of Twitter data, or more broadly User-generated Content. This is specifically relevant to our implementation using transformer based models, with
\citet{bertachillesheel} showing the reduction in the performance of BERT with the introduction of noise in the form of spelling mistakes and typos. Similar inaccuracies can be expected in the data set for the present task given the informal nature of Twitter, the distinctive character limit, and modification of words by user to indicate emotion.

In relation to overcoming the challenges of Twitter data specifically, \citet{ying-etal-2019-improving} leveraged a token pattern detector to obtain domain-specific features. This comprised a convolution network trained on annotated features derived from pre-processing for domain-specific features. These representations were concatenated with the $[$CLS$]$ token embedding from BERT, with the resulting model producing a statistically significant improvement on multi-label emotion classification over pure BERT \citep{ying-etal-2019-improving}.   

\subsection{Leveraging Metadata}
\label{sec:rl_metadata}
Social media's widespread use generates a vast amount of data, not just in text but additionally in metadata.  
In the body of work surrounding the classification of rumors on Twitter , metadata has been successfully leveraged to develop handcrafted features \citep{rumorli2019}. In the classification model produced by \citet{rumorli-etal-2019-},  metadata information specific to Twitter, such as  whether the account id is verified, if the profile includes a location or a description is concatenated together to form a input of "User Information" which is passed to a classifier along with textual and other features \citep{rumorli-etal-2019-}. This methodology of handcrafting features which provide supplementary information to a classifier is particularly related to the models in this work where we leverage a handcrafted feature derived from the text (Section~\ref{sec:meth}).

\section{Methodology}
\label{sec:meth}

The data for this work was provided by the Workshop on Noisy User-generated Text \citep{wnut2019} and consisted of 10,000 English tweets in relation to Covid-19. A 70/10/20 rate had been used to split the 10,000 tweets into training, validation and test sets. Each tweet had been labelled either as 'uninformative' or 'informative' by three independent annotators with an inter-annotator agreement score of Fleiss' Kappa at 0.818 \citep{wnut2019}. At the onset of this work, only the training and validation data had been provided.  The test set was retained as a holdout set for the Workshop to evaluate the performance of models, discussed at Section~\ref{sec:conclusion}.

\begin{figure}[!h]
    \centering
    \includegraphics[scale=0.45]{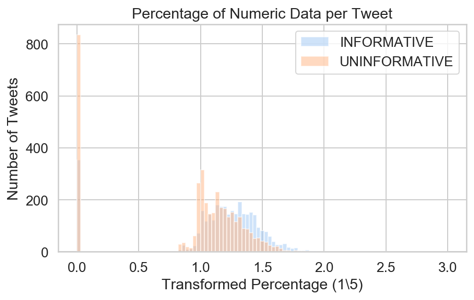}
    \caption{Probability of a character being numeric (power transformed $x^{\frac{1}{5}}$).} 
    \label{fig:graphnumeric}
\end{figure}

In an attempt to build upon the success of leveraging tweet metadata in a classification model (see Section~\ref{sec:rl_metadata}) an initial exploration of the distribution of metadata and text features between the ``uninformative'' and ``informative'' classes in the training dataset was conducted. This processes highlighted little differentiation between the classes, suggesting the distinction was more nuanced and conveyed within the circumstantial and contextual meanings. However, it was found that the higher the probability any given character in tweet was a numeric character increased the probability of it being ``informative''. We present a visual representation of this at Figure~\ref{fig:graphnumeric}.

By virtue of the competitive element, this task presented an incentive to find the best performing model. In order to have a basis for comparison two initial models were trained on the raw training data and evaluated against the validation data: \textbf{(1)} a Linear Support Vector Machine (SVM) using TF-IDF features, and \textbf{(2)} fine-tuned BERT\textsubscript{BASE}. As is the case with all the deep networks tested in this study, we test a range of hyperparameter values (details of which are available in the program code and model details released as part of this publication \footnote{\url{https://github.com/CalumPerrio/WNUT-2020}}), and test each combination against five different random seeds. The effect of random seed on BERT is emphasised in \citet{dodge2020finetuning} where varying only the random seed was shown to produce substantial improvements over previously published results. We present the highest performing results of these models at Table~\ref{table:baseline}. 

\begin{table}[!h]
\centering
\begin{tabular}{lrl}
\hline 
\textbf{Model} & \textbf{$F_1$ Score} & 
\\ \hline
SVM & 0.8155  \\
BERT & 0.9051 \\
\hline
RoBERTa (unprocessed) & 0.9101\\
RoBERTa (processed) & 0.9131\\
\hline
\end{tabular}
\caption{\label{table:baseline} Baseline results achieved by BERT and SVM, and results from RoBERTa. All tested aganist the validation data.}
\end{table}

The SVM model was chosen due to the established high accuracy this model can achieve on the task of text classification (\cite{Kadhim_2019}; \cite{svm2}).

We observe that a fine-tuned BERT model achieved a good performance against the validation data. The next logical progression was to experiment with a fine-tuned RoBERTa implementation, given that the robustly optimised pre-training this model employs improved performance over BERT \citep{RoBERTa}. We fine-tuned a RoBERTa\textsubscript{BASE} model, and the results of this are presented against the BERT and SVM baselines in Table~\ref{table:baseline}. We present two versions of the RoBERTa model: processed and unprocessed. The processed implementation draws inspiration from the pre-processing adopted by \citet{nikolov-radivchev-2019-nikolov}, in which we also segmented camel-cased hashtags into distinct tokens e.g. the token ``\#HashTag'' would become ``\#Hash'' and ``Tag''. Additionally, the two forms primarily used to refer to coronavirus: ``covid-19'' and ``coronavirus'' were parsed for in an extensive number of variations and standardised to "coronavirus" to ensure the token representation for this key term was consistent across inputs.

\begin{table}[!ht]
\centering
\begin{tabular}{lrl}
\hline 
\textbf{Model} & \textbf{False Negative} & \textbf{False Positive} 
\\ \hline
BERT  & 62.16\% & 41.81\% \\
RoBERTa & 56.6\% & 43.27\% \\
\hline
\end{tabular}
\caption{\label{table:intersection} Percentage of miss-classifications shared with the SVM. Visual representations are included in the Appendix.}
\end{table}

An error analysis was conducted after the development and testing of RoBERTa which explored the intersection of the false negative and false positive classifications from the SVM, BERT and the processed RoBERTa implementations. We present the most interesting finding from this in Table~\ref{table:intersection}: despite higher overall miss-classification, the SVM trained on TF-IDF features was able to correctly classify a significant proportion of miss-classifications from the fine-tuned BERT and RoBERTa models. This suggests, as in \citet{lim2020uob}, that incorporation of corpus level information could improve performance.

Additionally, the error analysis noted the BPE tokenization of ``coronavirus'' was out-of-vocabulary, splitting into the tokens ``Ġcoron, av, irus''. To this regard, post submission we have tested a promising implementation that instead replaces all forms referring to ``coronavirus'' to ``coronavirus disease''. With the intention of leveraging through self-attention the context of these terms as a disease. This is explored further in the future work.  

\subsection{Exploration of Improving Baseline Performances}
The error analysis and initial feature exploration motivated the experimentation of improving the performance of a pre-trained language model in the following three ensemble models. In all instances the RoBERTa processed implementation was used as the base pre-trained language model due to it achieving the highest $F_1$ score on the validation data in our initial experiments. 
 
Three ensemble models were experimented, presented broadly below and explored in details at Section~\ref{sec:meth:model_arch}, these were: \textbf{a)} RoBERTa together with a percentage metric of the probability of a character in the text being numeric, \textbf{b)} RoBERTa with TF-IDF features, and \textbf{c)} RoBERTa with both TF-IDF features and the percentage metric. 

\subsection{Model Architecture}
\label{sec:meth:model_arch}
For all three ensemble models, we concatenate the additional features to the final hidden layer for the $</s>$ token, which we use as an aggregate representation of the sequence. This vector then forms the input to the fully connected output layer. 

With regards to the TF-IDF features. A document-term matrix was constructed from a processed version of the training data, and a TF-IDF feature vector fit for the text for the equivalent tokenized input to the model. Processing involved removing punctuation, stopwords and emojis, and stemming. During testing, two combinations of maximum features in the document-term matrix were tested: 6000 and 9000. 

The probability metric was produced by first removing emojis from the text to prevent the influence of numbers in unicode. Then returning the probability of any character in the tweet being a digit, transformed as a (1,1) tensor. 

\section{Results and Analysis}
\label{sec:results}
We present the results of the three ensemble models in comparison to the processed RoBERTa model from Table~\ref{table:baseline} in Table~\ref{table:abalation}. 

\begin{table}[!h]
\centering
\begin{tabular}{lrl}
\hline 
\textbf{Model} & \textbf{$F_1$ Score}
\\ \hline
RoBERTa (baseline) & 0.9131 \\
\hline
RoBERTa+PROB & 0.9141 \\
RoBERTa+TFIDF & 0.9111\\
RoBERTa+PROB+TFIDF & 0.9151\\
\hline
\end{tabular}
\caption{\label{table:abalation} F1 score results of the three ensemble models presented against the baseline performance of RoBERTa.}
\end{table}

We observe that against the baseline RoBERTa model, inclusion of the hand-crated feature improved performance. However, the increase is small, potentially reflecting that the number of tweets in the data set for which this features is a useful indicator is equally small. We also observe that the inclusion of TF-IDF features on the validation data alone did not improve performance, however the inclusion of both the average metric and TF-IDF produced an improved result over the baseline and RoBERTa+PROB. The former result does not support our hypothesis of improving performance with the inclusion of TF-IDF features and is in contradiction to the latter result.

Between the RoBERTa+TFIDF model and RoBERTa+TFIDF+PROB models, the highest performances were found on models featuring 9000 and 6000 TF-IDF features respectively. It is notable that the total dimension of the stemmed document-term matrix was over 15,000 features. As such, due to memory constraints it was not possible to use a vector of this entire dimension. We submit an interesting experiment would be to parse for domain-specific expressions as in \citep{ying-etal-2019-improving} and standardise these to dictionary representation to reduce the number of TF-IDF features. Additionally, standardisation of these expressions across the corpus could potentially enable greater accuracy in calculating the Inverse Document Frequency, and by virtue of this better capture the amount of information a term provides.

In our submission to the evaluation stage we submit the RoBERTa+PROB+TFIDF ensemble model which achieved the highest performance of all models tested against the validation data (see Table~\ref{table:abalation}). Our results against the holdout test set are presented at Table~\ref{table:resutls} \footnote{Independently evaluated by organisers.}. Our ensemble model achieves an $F_1$ score of 0.8910, within two points of the highest performing team. 

\begin{table}[!h]
\centering
\begin{tabular}{lrl}
\hline 
\textbf{Rank} & \textbf{Team} & \textbf{F1 Score}
\\ \hline
1 & NutCracker & 0.9096 \\
2 & NLP\_North & 0.9096 \\
\textbf{...}\\
21 & cxp949 (this work) & 0.8910\\
\textbf{...}\\
55 & TMU-COVID19 & 0.5000\\
\hline

\hline
\end{tabular}
\caption{\label{table:resutls} F1 score of Final Model against the holdout test set. }
\end{table}

\subsection{Error Analysis}
\label{subsec:error_analysis}
\begin{figure}[!h]
    \centering
    \includegraphics[scale=0.45]{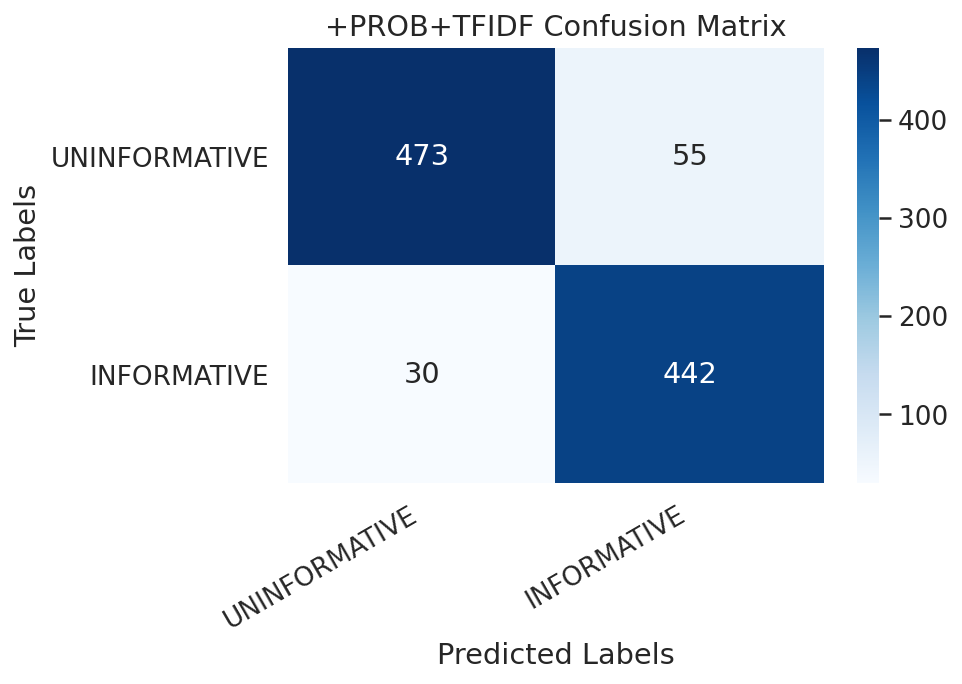}
    \includegraphics[scale=0.45]{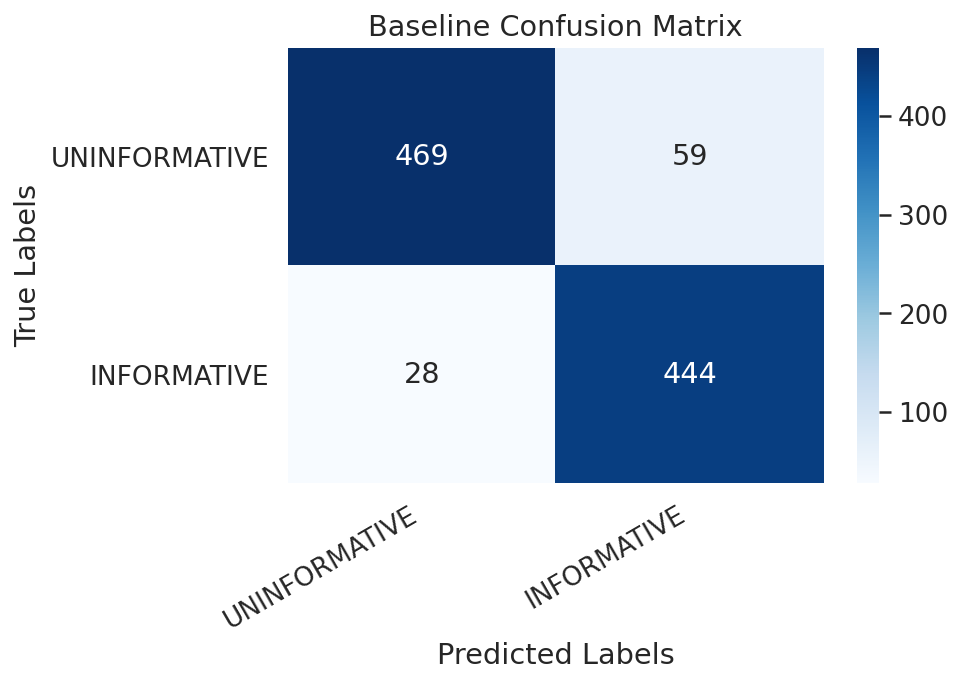}
    \caption{Confusion matrices of the RoBERTa+PROB+TFIDF ensemble model and RoBERTa baseline model respectively.} 
    \label{fig:confusion_matrix}
\end{figure}

We present at Figure~\ref{fig:confusion_matrix} the confusion matrices of the baseline RoBERTa model and RoBERTa+PROB+TFIDF ensemble model. We observe that inclusion of the handcrafted feature and TF-IDF features provided a net gain on true-positive "uninformative" classifications, and reduced the number of false-positive miss-classifications when compared with the baseline.  

We additionally explored the intersection of false-positive and false-negative miss-classifications between the two models. We observed that the intersect between the models comprised 87\% of the RoBERTa+PROB+TFIDF ensemble model's false-positive miss-classifications, and 77\% of the false-negative miss-classifications. We submit that this large intersection within the miss-classifications is indicative of a large proportion of tweets with which the RoBERTa model inherently struggles. \citet{ensembles} describe how the effectiveness of an ensemble model directly relies upon the accuracy and \textit{diversity} of the individual base learners, and to this regard we submit that more substantial improvements in performance require overcoming the limitations of the RoBERTa model in it's current form.


\section{Conclusions and Further Work}
\label{sec:conclusion}
We have explored with providing supplementary information to improve the performance of RoBERTa.
We have observed that the addition of a handcrafted feature improved performance of a pre-trained bidirectional transformer-based language model, suggesting that for text classification tasks and noisy data sets the inclusion of additional features that distinguish the classes can be beneficial. We intend to explore this concept further, beginning with conducting a statistical test to observe whether the difference between the presence of numeric data in informative tweets is significant.  

Additionally, we have experienced success with incorporating TF-IDF features with BERT to a degree, however we present that the size of the document-term matrix, the short length of tweets and domain-specific features of Twitter present potentially a challenge in utilising this optimally. 

As the global coronavirus pandemic continues to develop, the nature of what becomes ``informative'' information will likely develop also. Therefore, in the present task, an approach with a greater level of generalisation is arguably preferable. To this regard, initial testing of parsing the tweets and replacing the terms ``coronavirus'' and ``covid-19'' to ``coronavirus disease'', with a view to leveraging the existing embedding for "disease" to provide contextual information has shown promising results. The application of this new parsing objective presents an opportunity for future work. 

Furthermore, we observed in Section~\ref{sec:relatedwork} that conducting additional \textit{in-domain} pre-training was successfully utilised in relation to pre-training transformer-based models in a similar task. Additionally, in \citet{covidtwitterbert} the authors release a BERT model pre-trained on a data set of tweets in relation to the coronavirus that shows a 10-30\% marginal improvement on numerous classification data sets compared to BERT\textsubscript{LARGE}. Exploration of these two concepts may provide insight into improving the base RoBERTa model's performance (as discussed in Section~\ref{subsec:error_analysis}), and present further potential for future work and overcoming the challenges of a noisy data set centred around the topic of coronavirus.

\section*{Acknowledgements}

We would like to thank the NVIDIA Deep Learning Institute for the provision of AWS credits which we used to access GPU resources in this work. 

\bibliographystyle{acl_natbib}
\bibliography{emnlp2020}
\vfill
\pagebreak
\appendix
\section{Appendix}

\textbf{Pie Chart Representations of the Intersection between SVM, BERT and RoBERTa}

\vspace{1cm}

\includegraphics[scale=0.7]{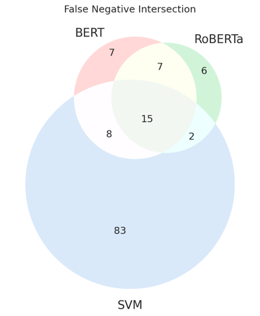}
\label{fig:fn_intersction}

\vspace{1cm}

\includegraphics[scale=0.7]{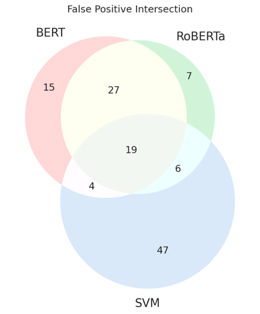}
\label{fig:fp_intersction}

\end{document}